\documentclass[lettersize,journal]{IEEEtran}
\usepackage{amsmath,amsfonts}
\usepackage[ruled,linesnumbered]{algorithm2e} 
\usepackage{amsmath}                         
\usepackage{bm}                              
\usepackage{amssymb} 
\usepackage{array}
\usepackage[caption=false,font=normalsize,labelfont=sf,textfont=sf]{subfig}
\usepackage{textcomp}
\usepackage{xspace}
\newcommand{\ie}{\textit{i.e.}\xspace}
\usepackage{stfloats}
\usepackage{url}
\usepackage{verbatim}
\usepackage[T1]{fontenc}
\usepackage[utf8]{inputenc}

\usepackage{booktabs}
\usepackage{multirow}
\usepackage{graphicx}

\usepackage{cite}
\begin{document}

\title{Dual Attention Guided Defense Against Malicious Edits}

\author{Jie Zhang,~\IEEEmembership{Member,~IEEE,}
\and
Shuai Dong~\IEEEmembership{,}
\and
Shiguang Shan~\IEEEmembership{Fellow,~IEEE,}
\and
Xilin Chen~\IEEEmembership{Fellow,~IEEE,}

\thanks{Jie Zhang, Shiguang Shan and Xilin Chen are with the State Key Laboratory of AI Safety, Institute of Computing Technology, Chinese Academy of Sciences (CAS), Beijing 100190, China, and also with the University of China Academy of Sciences, Beijing 100049, China (e-mail: zhangjie@ict.ac.cn; sgshan@ict.ac.cn; xlchen@ict.ac.cn).}
\thanks{Shuai Dong is with the School of Computer Science, China University of Geosciences, Wuhan 430074, China (e-mail: dongshuai\_iu@cug.edu.cn).}
}

\markboth{Journal of \LaTeX\ Class Files,~Vol.~14, No.~8, August~2021}%
{Shell \MakeLowercase{\textit{et al.}}: A Sample Article Using IEEEtran.cls for IEEE Journals}

\IEEEpubid{0000--0000/00\$00.00~\copyright~2021 IEEE}

\maketitle

\begin{abstract}
Recent progress in text-to-image diffusion models has transformed image editing via text prompts, yet this also introduces significant ethical challenges from potential misuse in creating deceptive or harmful content. While current defenses seek to mitigate this risk by embedding imperceptible perturbations, their effectiveness is limited against malicious tampering. To address this issue, we propose a Dual Attention-Guided Noise Perturbation (DANP) immunization method that adds imperceptible perturbations to disrupt the model's semantic understanding and generation process. DANP functions over multiple timesteps to manipulate both cross-attention maps and the noise prediction process, using a dynamic threshold to generate masks that identify text-relevant and irrelevant regions. It then reduces attention in relevant areas while increasing it in irrelevant ones, thereby misguides the edit towards incorrect regions and preserves the intended targets. Additionally, our method maximizes the discrepancy between the injected noise and the model's predicted noise to further interfere with the generation. By targeting both attention and noise prediction mechanisms, DANP exhibits impressive immunity against malicious edits, and extensive experiments confirm that our method achieves state-of-the-art performance.
\end{abstract}


\section{Introduction}
\IEEEPARstart{R}{ecent} advancements in diffusion models \cite{Ho2020DenoisingDP,SohlDickstein2015DeepUL,Song2020ScoreBasedGM} have significantly propelled generative modeling, especially in text-to-image generation. Models like DALLE2 \cite{Ramesh2022HierarchicalTI}, Imagen \cite{Saharia2022PhotorealisticTD}, and Stable Diffusion \cite{Rombach2021HighResolutionIS} leverage diffusion processes with natural language inputs to generate images that align with user-provided text, offering intuitive control over image generation. Beyond text-to-image tasks, diffusion models have been extended to image inpainting, editing, zero-shot classification, and open vocabulary segmentation \cite{Avrahami2021BlendedDF,Couairon2022DiffEditDS,Meng2021SDEditGI,Mokady2022NulltextIF,Wallace2022EDICTED,Wang2022ImagenEA,Xie2022SmartBrushTA,Xu2023OpenVocabularyPS,Zhu2023MovieFactoryAM}. Their progressive generation allows dynamic content adjustment, making them highly effective for image editing. However, the widespread use of diffusion models also raises concerns about potential negative impacts. Misuse of these technologies can lead to the creation of fake or manipulated images, deceiving the public, spreading misinformation, or manipulating opinion, thereby exacerbating societal trust issues. Additionally, privacy risks emerge when personal images are edited without consent, potentially resulting in privacy breaches or psychological harm. More seriously, these techniques can be used to produce harmful content, posing serious threats to social stability. Therefore, a thorough exploration of the security implications surrounding image editing technologies is essential to mitigate the risks of misuse. 

\IEEEpubidadjcol
In response to these challenges, two primary mitigation strategies have emerged, namely the reactive detection of manipulated content~\cite{Passos2022ARO,Naitali2023DeepfakeAG,Pei2024DeepfakeGA,wang2022lisiam,han2023fcd,yin2023dynamic} and the proactive immunization of original images against unauthorized editing~\cite{Aneja2021TAFIMTA,Ruiz2020DisruptingDA,Salman2023RaisingTC,Xue2023TowardEP}. Reactive detection operates post-facto by training classifiers to identify digital artifacts or inconsistencies left by generative models. However, this approach does not prevent the initial creation and spread of harmful content, as the damage may already be done by the time an image is flagged. In contrast, proactive immunization offers a more robust, pre-emptive defense by embedding imperceptible, adversarial perturbations into an original image before it is shared. The primary advantage of this proactive stance is its ability to disrupt the malicious editing process at its source, preventing harmful content from being successfully generated. Rather than merely identifying a fake, immunization aims to make its creation infeasible, thereby shifting the security burden from downstream detection to the point of content origin and empowering creators with direct control over their digital assets.

These immunization strategies primarily rely on adversarial attacks to disrupt the image editing process by introducing carefully crafted perturbations at various stages. While early immunization strategies are effective against GAN-based editing models~\cite{Aneja2021TAFIMTA,Ruiz2020DisruptingDA}, the advent of diffusion models and their robust denoising process necessitates new approaches. For diffusion models, initial defenses focus on adversarially attacking the VAE encoder or the entire generation process~\cite{Salman2023RaisingTC}. This has evolved into more refined strategies, such as disrupting semantic understanding by attacking cross-attention~\cite{Lo_2024_CVPR}, shifting latent representations to foil instruction-guided edits~\cite{Chen2023EditShieldPU}, preventing style mimicry with textural losses~\cite{Liang2023MistTI}, and enhancing attack efficiency and potency with techniques like score distillation sampling~\cite{Xue2023TowardEP} and targeted attacks~\cite{Zheng2023TargetedAI}. However, a key limitation of these approaches is their focus on isolated components (\textit{e.g.}, encoder, attention, or noise prediction) within the diffusion pipeline. This narrow scope creates a specific vulnerability that malicious actors could exploit, potentially limiting robustness.

In this paper, we propose Dual Attention-Guided Noise Perturbation (DANP), an immunization method engineered to generate images highly resistant to editing by diffusion-based models. Our approach is motivated by a key insight: effective editing relies on the two critical functions, \ie, the cross-attention mechanism and the stable denoising process, where the former acts as a semantic compass to localize regions based on text prompts and the latter serves as the generative engine that renders the edits. DAA manipulates the model’s semantic understanding across timesteps. First, it applies an adaptive dynamic threshold to cross-attention maps, generating precise masks that isolate text-relevant regions. Then, it conducts a dual-directional manipulation, which suppresses attention on relevant areas while intensifying it on irrelevant regions to render editing ineffective. Complementing this misdirection, the NBA directly attacks the generative engine by maximizing noise prediction discrepancies, \ie, the difference between injected image noise and the model’s predictions at key timesteps. This destabilizes the denoising trajectory, preventing convergence to clean, text-aligned outputs. Through synergistic action, DAA and NBA hinder both editing localization (``where'') and edit execution (``how''), enabling comprehensive, robust immunization.

It is noteworthy that while our DANP is closely related to SA~\cite{Lo_2024_CVPR} which also explore attacking attentions, it achieves more comprehensive protection through three key innovations. First, rather than applying a fixed threshold to sparse post-softmax attention maps as SA does, DANP first normalizes these maps to reveal a more distributed and informative representation of attention and then uses Kapur’s method to dynamically determine precise thresholds, yielding significantly more accurate masks than SA’s fixed approach. Second, while SA merely suppresses attention, DANP executes dual-directional manipulation: reducing attention in target regions while amplifying it in irrelevant areas to actively misdirect editing. Finally, DANP extends beyond SA’s singular focus on cross-attention via a dual-pronged strategy. DAA disrupts semantic alignment while the NBA simultaneously sabotages the denoising process, which ensures substantially stronger robustness against adaptive adversaries.

In summary, we make the following contributions:
\begin{itemize}
\item We propose the Dual Attention-Guided Noise Perturbation (DANP), a novel dual-pronged immunization framework that protects images by simultaneously disrupting the semantic understanding and generative process of diffusion models.
\item DANP integrates two core components: Dual Attention-guided Attack (DAA) which uses dynamically thresholded attention maps to attack attentions, \ie, suppressing relevant regions while amplifying irrelevant ones; and Noise-Based Attack (NBA), which injects noise to maximize discrepancy with model-predicted noise, thereby degrading the denoising process.
\item Extensive experiments demonstrate that DANP achieves state-of-the-art performance, offering superior robustness and protection against malicious edits compared to existing methods.
\end{itemize}

\section{Related work}
\subsection{Image Editing}

Image editing aims to manipulate the content of a given image based on a provided editing prompt. Early works predominantly rely on generative adversarial networks (GANs) for this task \cite{zhu2017unpaired,karras2019style,shen2020interpreting}. Notable methods like CycleGAN \cite{zhu2017unpaired} and StyleGAN \cite{karras2019style} achieve impressive results in image generation and editing. However, GAN-based methods often struggle with issues related to model generalization and image quality due to inherent limitations in their architectures. Recently, diffusion models \cite{ho2020denoising,dhariwal2021diffusion} have made remarkable progress in text-to-image generation, owing to their strong generative capabilities and stable training processes. Building upon this, diffusion models have been extended to text-guided image editing. As one of the earliest diffusion-based image editing methods, SDEdit \cite{meng2021sdedit} utilizes stochastic differential equations to progressively refine an input based on a textual prompt, enabling nuanced and controllable edits. Following this, Avrahami \textit{et al.} \cite{avrahami2022blended} proposes Blended Diffusion, which integrates textual guidance into the diffusion process to modify specific regions of an image while preserving unedited content. DiffusionCLIP \cite{kim2021diffusionclip} further advances the field by combining diffusion models with CLIP embeddings, enabling text-guided image manipulation without requiring paired text-image data for training. Hertz \textit{et al.} \cite{hertz2022prompt} introduces cross-attention control mechanisms to improve alignment between the edited image and the textual prompt, allowing for more precise and flexible editing. More recently, Zhang \textit{et al.} \cite{zhang2023controlnet} adds conditional control to text-to-image diffusion models by incorporating additional control signals like edge maps or segmentation masks, leading to precise control over the generated content.

\subsection{Image Immunization}
The rise of diffusion-based image editing models has made malicious tampering more accessible, underscoring the need to protect images from unauthorized edits. Image immunization addresses this by adding imperceptible perturbations that prevent editing models from modifying images based on specific instructions or text prompts. While early research focuses on GAN-based models~\cite{Ruiz2020DisruptingDA, Aneja2021TAFIMTA}, the unique structure of diffusion models necessitate new strategies. Influential approaches like PG~\cite{Salman2023RaisingTC} establish the core strategy of attacking the model's latent space to disrupt editing and style mimicry. This has been extended by methods like Mist~\cite{Liang2023MistTI}, which combines this latent-space attack with a semantic loss, and ED~\cite{Chen2023EditShieldPU}, which specifically foils instruction-guided models by shifting latent representations. Other works have broadened the attack surface with more efficient and potent methods. These include score distillation sampling~\cite{Xue2023TowardEP}, which improves efficiency by approximating gradient calculations, and targeted attacks against model customization~\cite{Zheng2023TargetedAI}, which minimize the distance between the model's output and a predefined target. The work most closely related to ours, SA~\cite{Lo_2024_CVPR}, specifically targets the cross-attention mechanism to impair semantic understanding by suppressing attention in text-relevant regions. However, this singular focus on suppressing attention neglects the potential of manipulating irrelevant regions to achieve a more comprehensive defense, thereby limiting its effectiveness. To overcome this limitation, we propose a novel method that achieves superior immunization by manipulating attention in both relevant and irrelevant areas, representing a significant advancement in the field.

\section{Preliminaries}

This section outlines the foundational concepts essential for understanding our proposed method, DANP. We first review the operational principles of conditional diffusion models, then detail the cross-attention mechanism that facilitates text-guided editing, and finally introduce the core principles of image immunization.

\subsection{Conditional Diffusion Models}
Diffusion models~\cite{ho2020denoising,dhariwal2021diffusion} generate images by iteratively refining random noise through a denoising process. Starting from pure noise $\mathbf{x}_T$, the model predicts less noisy versions $\mathbf{x}_{t-1}$ at each timestep $t$, guided by a condition $\mathbf{c}$ such as a text prompt:
\begin{equation}
    \mathbf{x}_{t-1} = \frac{1}{\sqrt{\alpha_t}} \left( \mathbf{x}_t - \frac{1 - \alpha_t}{\sqrt{1 - \bar{\alpha}_t}} \boldsymbol{\epsilon}_{\boldsymbol{\theta}}(\mathbf{x}_t, t, \mathbf{c}) \right) + \sigma_t \boldsymbol{\epsilon},
    \label{eq:reverse_process}
\end{equation}
where $\boldsymbol{\epsilon}_{\boldsymbol{\theta}}$ is the noise prediction model parameterized by $\boldsymbol{\theta}$, $\alpha_t$ and $\bar{\alpha}_t$ represent the scalar noise scheduling parameters, $\sigma_t$ is the standard deviation, and $\boldsymbol{\epsilon} \sim \mathcal{N}(0, \mathbf{I})$ is a random noise vector.

The model is trained to minimize the noise prediction error, which is given by:
\begin{equation}
    \mathcal{L}(\boldsymbol{\theta}) = \mathbb{E}_{\mathbf{x}_0, \mathbf{c}, \boldsymbol{\epsilon}, t} \left[ \left\| \boldsymbol{\epsilon} - \boldsymbol{\epsilon}_{\boldsymbol{\theta}}(\mathbf{x}_t, t, \mathbf{c}) \right\|^2 \right],
\end{equation}
where $\mathbf{x}_t = \sqrt{\bar{\alpha}_t} \mathbf{x}_0 + \sqrt{1 - \bar{\alpha}_t} \boldsymbol{\epsilon}$.

\subsection{Cross-Attention Mechanism in Diffusion Models}
Given a text prompt $\mathbf{c}$ and an image $\mathbf{x}$, the cross-attention mechanism is crucial for aligning textual and visual features. The attention map (a matrix) at the $l$-th intermediate block of the U-Net is defined as:
\begin{equation}
\mathbf{A}_l(\mathbf{x}, \varphi(\mathbf{c})) = \text{softmax}\left(\frac{\mathbf{Q}_l \mathbf{K}_l^\top}{\sqrt{d_k}}\right),
\end{equation}
where $\varphi(\cdot)$ is the text encoder, $\mathbf{Q}_l$ (queries) are matrices derived from image features, and $\mathbf{K}_l$ (keys) are matrices from text embeddings. Since the U-Net architecture~\cite{DBLP:journals/corr/RonnebergerFB15, Cao2021SwinUnetUP} generates multiple attention maps, they are typically aggregated:
\begin{equation}
	\mathbf{Att}(\mathbf{x}, \varphi(\mathbf{c})) = \frac{1}{L} \sum_{l=1}^L \text{Upsample}(\mathbf{A}_l(\mathbf{x}, \varphi(\mathbf{c}))),
    \label{eq:agg_attn}
\end{equation}
where $L$ is the number of U-Net blocks. This aggregated attention map, $\mathbf{Att}(\mathbf{x}, \varphi(\mathbf{c}))$, provides a spatial understanding of where textual concepts are located within the image.

\subsection{Image Immunization}

The objective of image immunization is to craft an adversarial example $\mathbf{x}_{\text{adv}} = \mathbf{x}_0 + \boldsymbol{\delta}$ by adding an imperceptible perturbation $\boldsymbol{\delta}$ to a source image $\mathbf{x}_0$. This perturbation is designed to ensure that the output of an editing model $f_{\boldsymbol{\theta}}(\mathbf{x}_{\text{adv}}, \mathbf{c})$ is substantially altered compared to the edit of the original image, $f_{\boldsymbol{\theta}}(\mathbf{x}_0, \mathbf{c})$.

From an adversarial perspective~\cite{Goodfellow2014ExplainingAH, Madry2017TowardsDL}, this can be framed as an optimization problem where the optimal perturbation $\boldsymbol{\delta}^*$ is found by maximizing the dissimilarity between the two outputs, subject to a constraint on the perturbation's magnitude, typically an $L_p$-norm ball where $\|\boldsymbol{\delta}\|_p \leq \epsilon$:
\begin{equation}
\boldsymbol{\delta}^* = \underset{\|\boldsymbol{\delta}\|_p \leq \epsilon}{\arg\max}\, \mathcal{L}\left( f_{\boldsymbol{\theta}}(\mathbf{x}_0 + \boldsymbol{\delta}, \mathbf{c}), f_{\boldsymbol{\theta}}(\mathbf{x}_0, \mathbf{c}) \right).
\label{eq:immunization_objective_alt}
\end{equation}
In this formulation, $\mathcal{L}$ represents a differentiable metric, such as the $\ell_2$ distance, quantifying the dissimilarity between the outputs. To solve for $\boldsymbol{\delta}^*$, iterative algorithms like Projected Gradient Descent (PGD)~\cite{Madry2017TowardsDL} are commonly employed. At each iteration $k$, the perturbation is updated according to the rule:
\begin{equation}
    \boldsymbol{\delta}_{k+1} = \Pi_{\|\cdot\|_p \leq \epsilon} \left( \boldsymbol{\delta}_{k} + \alpha \cdot \mathbf{g}_k \right),
    \label{eq:pgd_update_immunization_alt}
\end{equation}
where $\alpha$ is the step size, $\mathbf{g}_k$ is the normalized gradient direction (= $\text{sign}(\nabla_{\boldsymbol{\delta}_{k}} \mathcal{L})$), and $\Pi_{\|\cdot\|_p \leq \epsilon}(\cdot)$ projects the perturbation onto the $L_p$-ball.

\section{Method}
\subsection{Overview}
\begin{figure*}[t]
    \centering
    \includegraphics[width=1\linewidth]{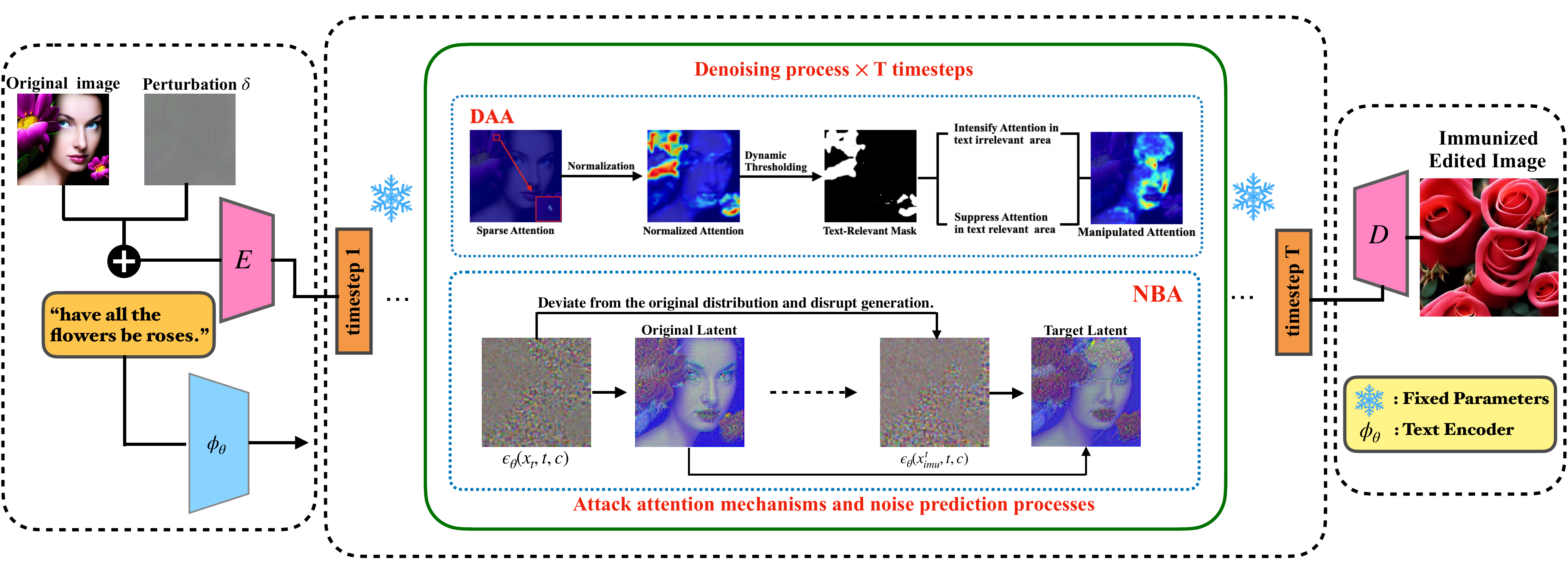}
    \caption{Overview of our DANP. The goal of the image editing model is to successfully modify the image according to a textual prompt, generating a realistic output with a diffusion model that matches the prompt. DANP introduces imperceptible perturbations that induce erroneous noise predictions in the editing model and disrupt cross attention maps by suppressing attention scores in text-relevant regions while elevating those in text-irrelevant areas, thereby generating images that fail to implement the intended edits.}
    \label{fig:framework}
\end{figure*}

In this work, we propose the Dual Attention-Guided Noise Perturbation (DANP), an immunization method that protects images from unauthorized edits by simultaneously disrupting the semantic and generative processes of diffusion models. The overall framework of our method is illustrated in Fig.~\ref{fig:framework}. First, we formulate a dual-pronged attack strategy targeting both the cross-attention mechanism for localizatio, and the denoising process for generation. We then iteratively optimize an imperceptible perturbation $\boldsymbol{\delta}$ using two synergistic components: a Dual Attention-guided Attack (DAA) and a Noise-Based Attack (NBA). The DAA objective manipulates attention maps to misdirect the model's focus away from the intended edit regions and towards irrelevant areas. Simultaneously, the NBA objective corrupts the denoising path by maximizing the error in the model's noise prediction. Once optimized, this perturbation is added to the original image to create an immunized version. Attempting to edit this immunized image causes the embedded DANP perturbation to disrupt the diffusion model, preventing it from both localizing and rendering the edit. This failure blocks any successful modification, ensuring the original image's integrity.

\subsection{Dual Attention-Guided Attack}
The Dual Attention-Guided Attack (DAA) is the first core component of our DANP framework, engineered to disrupt the semantic localization of diffusion-based editing models. It is achieved by strategically manipulating the cross-attention mechanism, which is essential for the model to understand where to apply a text-guided edit. Our approach not only shields the intended edit regions but also actively misguides the model's focus toward text-irrelevant areas, thereby causing the editing process to fail.

\begin{figure}[t]
    \centering
    \includegraphics[width=1\linewidth]{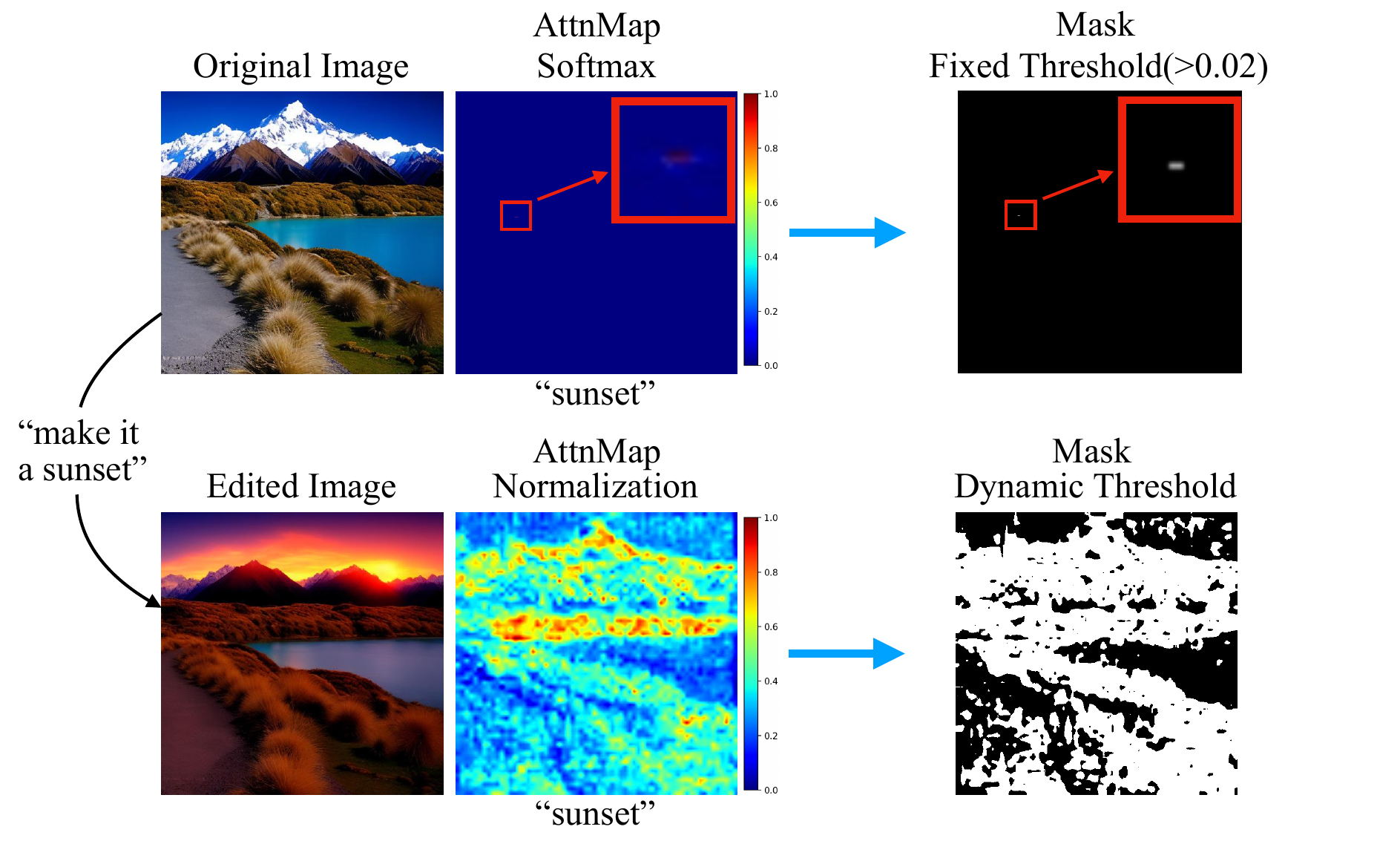}
    \caption{Comparison of fixed versus dynamic thresholding for attention mask generation. The top row demonstrates the limitation of applying a fixed threshold greater than 0.02 to the sparse post-softmax attention map, which only identifies a minimal and incomplete region. In contrast, the bottom row showcases our approach: first, the attention map is normalized to create a more effective and distributed representation. Subsequently, a dynamic threshold determined via Kapur's method, is applied to generate a more comprehensive and precise mask, accurately identifying regions relevant to the word `sunset'.}
    \label{fig:threshold_comparison}
\end{figure}

A critical step in manipulating attention is accurately identifying text-relevant regions. Prior work such as SA~\cite{Lo_2024_CVPR} attempts this by applying a fixed threshold directly to the raw, post-softmax attention maps. However, these maps are often extremely sparse, with attention concentrated in a few high-value peaks. As illustrated in the top row of Fig.~\ref{fig:threshold_comparison}, applying a fixed threshold (\text{e.g.}, $>0.02$) to such a sparse map generates an inadequate mask that fails to capture the full semantic region corresponding to the word `sunset'. It is inherently difficult to define a universally effective threshold for generating reliable masks across different cases.

To overcome this limitation, our DAA introduces a more robust two-stage process for mask generation. First, we normalize the aggregated attention map to a range of $[0, 1]$. This crucial step transforms the sparse map into a denser, more continuous representation that better reflects the relative importance of all image regions. Second, we apply a dynamic thresholding technique, specifically \textit{Kapur's entropy method}. Kapur's method is an automatic image thresholding approach that separates an image into foreground and background classes by maximizing the sum of their Shannon entropies, thereby achieving maximum information separability. Formally, given a single-channel image (e.g., an attention map) with $L$ intensity levels, its histogram is normalized to form a probability distribution $p_0, p_1, \dots, p_{L-1}$. For a threshold $\tau$, pixels are split into two classes: Class 0 (background) $[0, \dots, \tau]$ and Class 1 (foreground) $[\tau+1, \dots, L-1]$, with probabilities $P_0(\tau)$ and $P_1(\tau)$. The class entropies are
\begin{align}
    H_0(\tau) &= -\sum_{i=0}^{\tau} \frac{p_i}{P_0(\tau)} \log\left(\frac{p_i}{P_0(\tau)}\right), \\
    H_1(\tau) &= -\sum_{i=\tau+1}^{L-1} \frac{p_i}{P_1(\tau)} \log\left(\frac{p_i}{P_1(\tau)}\right),
\end{align}
and the optimal threshold $\tau^*$ is obtained by
\begin{equation}
    \tau^* = \underset{0 \le \tau < L}{\arg\max} \,[ H_0(\tau) + H_1(\tau) ].
\end{equation}

In our experiment, this procedure is applied to the normalized attention map at each timestep $t$ to compute the adaptive threshold $\tau_t$, enabling robust separation of text-relevant and irrelevant regions. As demonstrated in the bottom row of Fig.~\ref{fig:threshold_comparison}, this approach yields a far more comprehensive and accurate mask. The mask $\mathbf{M}_t$ generated at each timestep $t$ is formally defined by
\begin{equation}
	\mathbf{M}_t = \mathbb{I}(\mathcal{N}(\text{Att}(\mathbf{x}_t^{\text{imu}}, \varphi(\mathbf{c}))) > \tau_t) \label{eq:mask_gen}
\end{equation}
where $\mathbf{x}_t^{\text{imu}}$ is the perturbed image at timestep $t$, $\mathcal{N}(\cdot)$ is the normalization function, $\text{Att}(\cdot)$ is the aggregated attention map from a previously defined equation, $\tau_t$ is the dynamically computed threshold, and $\mathbb{I}(\cdot)$ is the indicator function.

\begin{figure}[t]
    \centering
    \includegraphics[width=1\linewidth]{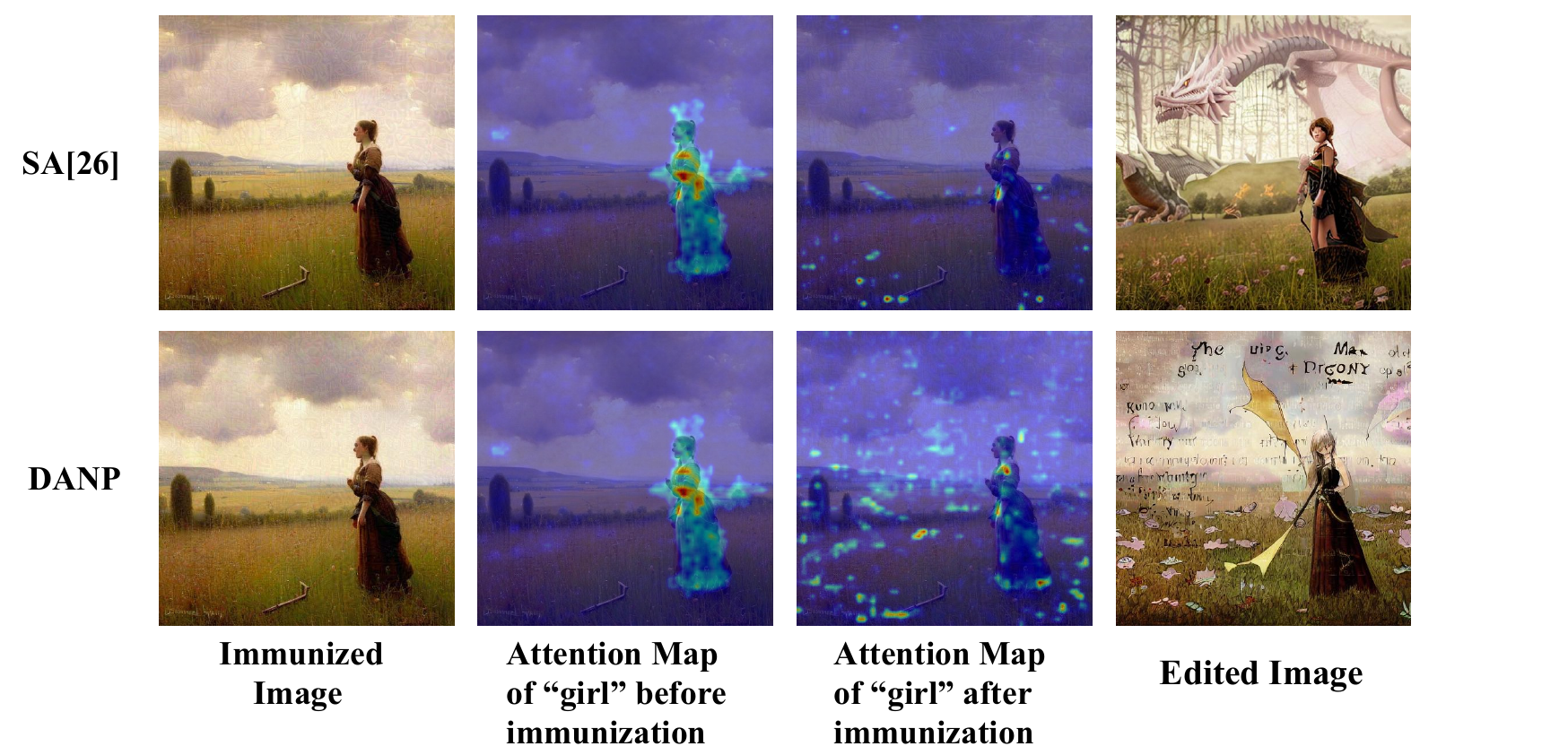}
    \caption{Visual comparison of attention map manipulation and immunization effects between our DANP and SA~\cite{Lo_2024_CVPR}. The editing prompt is `make the girl a giant dragon'. Unlike SA~\cite{Lo_2024_CVPR} that solely suppresses text-related attention, our method both suppresses text-related regions and amplifies irrelevant areas, diverting model editing to erroneous regions.}
    \label{fig:sa_vs_danp}
\end{figure}

With an accurate mask $\mathbf{M}_t$, we then introduce the central mechanism of DAA, which is a dual-directional manipulation of attention. Unlike methods that only suppress attention~\cite{Lo_2024_CVPR}, we recognize the critical importance of text-irrelevant regions in actively confusing the model. Consequently, our DAA loss is designed not only to suppress attention in text-relevant areas (where $\mathbf{M}_t = 1$) but also to simultaneously elevate it in text-irrelevant ones (where $\mathbf{M}_t = 0$). This dual-action approach is visualized in Fig.~\ref{fig:sa_vs_danp}, where DANP creates a grossly erroneous attention map, in contrast to the simple suppression seen in SA~\cite{Lo_2024_CVPR}. This forces the model to divert its generative power to incorrect regions. The objective is formulated using the contrastive loss function $\mathcal{L}_{\text{DAA}}$
\begin{equation}
\begin{split}
    \mathcal{L}_{\text{DAA}} ={}& \left\| \text{Att}(\mathbf{x}_t^{\text{imu}}, \varphi(\mathbf{c})) \odot \mathbf{M}_t \right\|_F^2 \\
                               & - \lambda_{\text{daa}} \left\| \text{Att}(\mathbf{x}_t^{\text{imu}}, \varphi(\mathbf{c})) \odot (1 - \mathbf{M}_t) \right\|_F^2
\end{split}
\label{eq:daa_loss}
\end{equation}
where $\odot$ denotes the element-wise product, $\|\cdot\|_F^2$ is the squared Frobenius norm, and $\lambda_{\text{daa}}$ is a hyperparameter to balance the two terms. By minimizing this loss, we actively divert the model's attention from relevant regions to erroneous ones, causing a systematic failure of the target edit.

\begin{algorithm}[t]
\caption{DANP Immunization Algorithm}
\label{alg:danp}
\KwIn{Source image $\mathbf{x}_0$, editing prompt embedding $\mathbf{c}$, perturbation budget $\gamma$, step size $\alpha$, number of iterations $N$, set of diffusion timesteps $\mathcal{T}$, loss weight $\lambda_{\text{nba}}$}
\KwOut{Immunized image $\mathbf{x}_{\text{imu}}$}
\BlankLine
$\boldsymbol{\delta} \leftarrow \mathbf{0}$\;
\For{$n = 1$ \KwTo $N$}{
    $\mathbf{g}_{\text{total}} \leftarrow \mathbf{0}$\;
    $\mathbf{x}_{\text{imu}} \leftarrow \mathbf{x}_0 + \boldsymbol{\delta}$\;
    \For{$t$ \textbf{in} $\mathcal{T}$}{
        Sample noise $\boldsymbol{\epsilon} \sim \mathcal{N}(\mathbf{0}, \mathbf{I})$\;
        $\mathbf{x}_t \leftarrow \sqrt{\bar{\alpha}_t} \mathbf{x}_0 + \sqrt{1 - \bar{\alpha}_t} \boldsymbol{\epsilon}$\;
        $\mathbf{x}_t^{\text{imu}} \leftarrow \sqrt{\bar{\alpha}_t} \mathbf{x}_{\text{imu}} + \sqrt{1 - \bar{\alpha}_t} \boldsymbol{\epsilon}$\;
        
        $\mathcal{L}_{\text{total}} \leftarrow \mathcal{L}_{\text{DAA}}(\mathbf{x}_t^{\text{imu}}, \mathbf{c}) + \lambda_{\text{nba}} \mathcal{L}_{\text{NBA}}(\mathbf{x}_t, \mathbf{x}_t^{\text{imu}}, \mathbf{c})$\;
        
        $\mathbf{g}_{\text{total}} \leftarrow \mathbf{g}_{\text{total}} + \nabla_{\mathbf{x}_{\text{imu}}} \mathcal{L}_{\text{total}}$\;
    }
    $\mathbf{g}_{\text{total}} \leftarrow \mathbf{g}_{\text{total}} / |\mathcal{T}|$\;
    
    $\boldsymbol{\delta} \leftarrow \boldsymbol{\delta} - \alpha \cdot \text{sign}(\mathbf{g}_{\text{total}})$\;
    
    $\boldsymbol{\delta} \leftarrow \text{clip}(\boldsymbol{\delta}, -\gamma, \gamma)$\;
}
$\mathbf{x}_{\text{imu}} \leftarrow \mathbf{x}_0 + \boldsymbol{\delta}$\;
\Return $\mathbf{x}_{\text{imu}}$\;
\end{algorithm}

\subsection{Noise-Based Attack}
While the DAA disrupts the model’s semantic guidance, the Noise-Based Attack (NBA) complements this by directly targeting the diffusion model’s core generative engine. Since image editing relies on iterative noise prediction and removal during reverse diffusion to render edits, the NBA corrupts this refinement process by systematically degrading generative capability.

Specifically, at each selected timestep $t$, our NBA aims to maximize the discrepancy between the noise that the model would predict for the original, unperturbed image $\mathbf{x}_t$ and the noise it predicts for the immunized $\mathbf{x}_t^{\text{imu}}$. By pushing these two predictions as far apart as possible, we effectively derail the denoising trajectory for the immunized image, forcing it onto an erroneous generation path that deviates significantly from the intended edit. This objective is formulated by minimizing the following loss function, which corresponds to maximizing the squared $L_2$-distance between the two noise predictions:
\begin{equation}
	\mathcal{L}_{\text{NBA}} = - \left\| \boldsymbol{\epsilon}_\theta(\mathbf{x}_t, t, \mathbf{c}) - \boldsymbol{\epsilon}_\theta(\mathbf{x}_t^{\text{imu}}, t, \mathbf{c}) \right\|_2^2
	\label{eq:nba_loss}
\end{equation}
where $\boldsymbol{\epsilon}_\theta$ is the noise prediction model, $\mathbf{x}_t$ is the noisy latent of the original image, and $\mathbf{x}_t^{\text{imu}}$ is the noisy latent of the immunized image. Minimizing $\mathcal{L}_{\text{NBA}}$ fundamentally corrupts the refinement guidance for immunized images, preventing convergence to coherent, text-aligned outputs.

\subsection{Overall Objective and Optimization}
Our Dual Attention-Guided Noise Perturbation (DANP) framework integrates the DAA and NBA components into a unified optimization objective. The goal is to iteratively craft an imperceptible perturbation $\boldsymbol{\delta}$ that simultaneously disrupts both the model's semantic understanding and its generative process.

The overall loss function $\mathcal{L}_{\text{total}}$ is defined as a weighted sum of the losses from each component:
\begin{equation}
	\mathcal{L}_{\text{total}} = \mathcal{L}_{\text{DAA}} + \lambda_{\text{nba}}\mathcal{L}_{\text{NBA}}
	\label{eq:total_loss}
\end{equation}
where $\mathcal{L}_{\text{DAA}}$ is the dual attention-guided loss from Eq.~(\ref{eq:daa_loss}), $\mathcal{L}_{\text{NBA}}$ is the noise-based loss from Eq.~(\ref{eq:nba_loss}), and $\lambda_{\text{nba}}$ is a hyperparameter that balances their contributions. By minimizing this combined loss, we compel the editing model to misinterpret where to apply the edit while simultaneously corrupting its ability to render the changes correctly.

To find the optimal perturbation, we minimize $\mathcal{L}_{\text{total}}$ using the iterative method detailed in Algorithm~\ref{alg:danp}. The optimization is performed by averaging gradients calculated over a set of diffusion timesteps $\mathcal{T}$, ensuring the attack is effective across different stages of the generation process. Furthermore, to maintain the visual fidelity of the original image, we constrain the perturbation to be imperceptible. This is enforced by ensuring the final immunized image $\mathbf{x}_{\text{imu}}$ remains within an $L_\infty$-norm ball of radius $\gamma$ around the source image $\mathbf{x}_0$, such that $\|\mathbf{x}_{\text{imu}} - \mathbf{x}_0\|_\infty \le \gamma$.

By targeting the fundamental mechanisms of both attention and denoising, our DANP framework is designed for a stronger and more robust immunization effect. This dual-pronged strategy ensures a more comprehensive disruption of the editing process, as it compromises both the model's ability to understand where to edit and its capacity to properly generate how to perform that edit.

\section{Experiments}
\subsection{Experimental Setup}

\paragraph{Target Models and Dataset}
Our experiments are conducted on several open-source models, including \texttt{StableDiffusion-v1-4}~\cite{Rombach2021HighResolutionIS}, \texttt{HQ-Edit}~\cite{Hui2024HQEditAH}, and \texttt{Instructpix2pix}~\cite{Brooks2022InstructPix2PixLT}, all accessible via Hugging Face~\cite{huggingface}. In the absence of a standard benchmark, we curate a diverse dataset of 200 images from the \texttt{Instructpix2pix-clip-filtered}~\cite{instructpix2pix-clip-filtered} collection, encompassing portraits, landscapes, and various artworks. For each image, the immunization perturbation is generated using the original editing prompt associated with that image in the dataset.

\paragraph{Evaluation on Unseen Prompts}
To assess robustness against unknown attacks, we evaluate all methods on unseen prompts. For each of the 200 source images, we manually craft five new, semantically distinct editing prompts. This results in a test set of 1,000 unique image-prompt pairs, allowing for a thorough evaluation of each method's generalization capabilities.

\paragraph{General Settings and Baselines}
To ensure a fair comparison, all methods are constrained by an $L_\infty$ perturbation budget of $\gamma = 0.03$ and are optimized for $N=100$ iterations. For methods involving diffusion timesteps, we uniformly sample a set of $|\mathcal{T}|=10$ timesteps to attack. We configure the baseline methods according to their original papers. 


\paragraph{Settings of DANP}
For our proposed DANP method, we set the hyperparameters as follows. The weight for the attention elevation term in the DAA loss (Eq.~(\ref{eq:daa_loss})) is set to $\lambda_{\text{daa}} = 1.0$, and the balancing weight for the NBA loss in the total objective (Eq.~(\ref{eq:total_loss})) is set to $\lambda_{\text{nba}} = 1.0$. Additionally, for the dynamic thresholding process within the DAA component, the number of intensity levels for Kapur's entropy method is set to $L=128$. Detailed ablation studies on these hyperparameters are provided in the subsequent sections.

\subsection{Evaluation Metrics}
To quantitatively evaluate the immunization effectiveness of each method, we employ a suite of metrics that measure the dissimilarity between the original edited image and the actual output produced from editing an immunized image. For all metrics, the goal is to quantify a large deviation from the original edited image, which signifies a successful defense.

We report several metrics traditionally used for measuring similarity, where \textit{lower values indicate a more successful defense}. These include the Peak Signal-to-Noise Ratio (PSNR), which measures pixel-wise mean squared error; the Structural Similarity Index (SSIM)~\cite{1284395}, which compares structural information, luminance, and contrast; the Feature Similarity Index (FSIM)~\cite{5705575}, which compares low-level image features; and the Visual Information Fidelity in the pixel domain (VIFp)~\cite{1576816}, which quantifies differences in preserved visual information. 
Furthermore, we utilize perception-based distance metric where \textit{higher values indicate stronger defense}. The Learned Perceptual Image Patch Similarity (LPIPS)~\cite{Zhang2018TheUE} leverages deep features to calculate the perceptual distance between the editing result of original image and immunized image.

\subsection{Main Results}
We compare our method with a suite of recent baseline approaches, including ACE~\cite{Zheng2023TargetedAI}, ED~\cite{Chen2023EditShieldPU}, MIST~\cite{Liang2023MistTI}, PGD, PGE~\cite{Salman2023RaisingTC}, SDS~\cite{Xue2023TowardEP}, and SA~\cite{Lo_2024_CVPR}. These experiments, performed across the three distinct and widely-used editing models of \texttt{Instructpix2pix}, \texttt{StableDiffusion-v1-4}, and \texttt{HQ-Edit}, consistently show that DANP achieves significantly superior performance. The following sections detail these results for each model.

\begin{table*}[t!]
\centering
\caption{Quantitative performance comparison of various immunization methods on the \texttt{Instructpix2pix} model, evaluated under both original and unseen prompt scenarios.}
\label{tab:quantitative_comparison_fid}
\resizebox{\textwidth}{!}{%
\begin{tabular}{l ccccc ccccc }
\toprule
Method & \multicolumn{5}{c}{Original Prompt} & \multicolumn{5}{c}{Unseen Prompts} \\
\cmidrule(lr){2-6} \cmidrule(lr){7-11}
& PSNR $\downarrow$ & SSIM $\downarrow$ & LPIPS $\uparrow$ & VIFp $\downarrow$ & FSIM $\downarrow$ & PSNR $\downarrow$ & SSIM $\downarrow$ & LPIPS $\uparrow$ & VIFp $\downarrow$ & FSIM $\downarrow$ \\
\midrule
ACE~\cite{Zheng2023TargetedAI}    & 16.89 & 0.6210 & 0.4307 & 0.2054 & 0.8019 & 16.19 & 0.5798 & 0.4343 & 0.1770 & 0.7799 \\
SDS~\cite{Xue2023TowardEP}    & 15.45 & 0.5951 & 0.4791 & 0.1776 & 0.7814 & 15.78 & 0.5691 & 0.4496 & 0.1763 & 0.7738 \\
ED~\cite{Chen2023EditShieldPU}     & 18.71 & 0.7157 & 0.3152 & 0.3101 & 0.8448 & 17.12 & 0.6235 & 0.3823 & 0.2243 & 0.7999 \\
MIST~\cite{Liang2023MistTI}   & 16.12 & 0.6072 & 0.4566 & 0.2078 & 0.7990 & 16.38 & 0.5832 & 0.4367 & 0.1877 & 0.7815 \\
PGD~\cite{Salman2023RaisingTC}    & 16.28 & 0.6221 & 0.4225 & 0.2115 & 0.7956 & 15.90 & 0.5763 & 0.4366 & 0.1787 & 0.7744 \\
PGE~\cite{Salman2023RaisingTC}    & 15.75 & 0.5586 & 0.4652 & 0.1875 & 0.7691 & 15.65 & 0.5565 & 0.4487 & 0.1668 & 0.7677 \\
SA~\cite{Lo_2024_CVPR}     & 16.17 & 0.6045 & 0.4626 & 0.1986 & 0.7973 & 15.97 & 0.5748 & 0.4398 & 0.1788 & 0.7786 \\
\textbf{DANP} & \textbf{14.67} & \textbf{0.5487} & \textbf{0.4861} & \textbf{0.1764} & \textbf{0.7519} & \textbf{14.99} & \textbf{0.5498} & \textbf{0.4651} & \textbf{0.1520} & \textbf{0.7611} \\
\bottomrule
\end{tabular}%
}
\end{table*}

\begin{figure*}[t]
    \centering
    \includegraphics[width=1\linewidth]{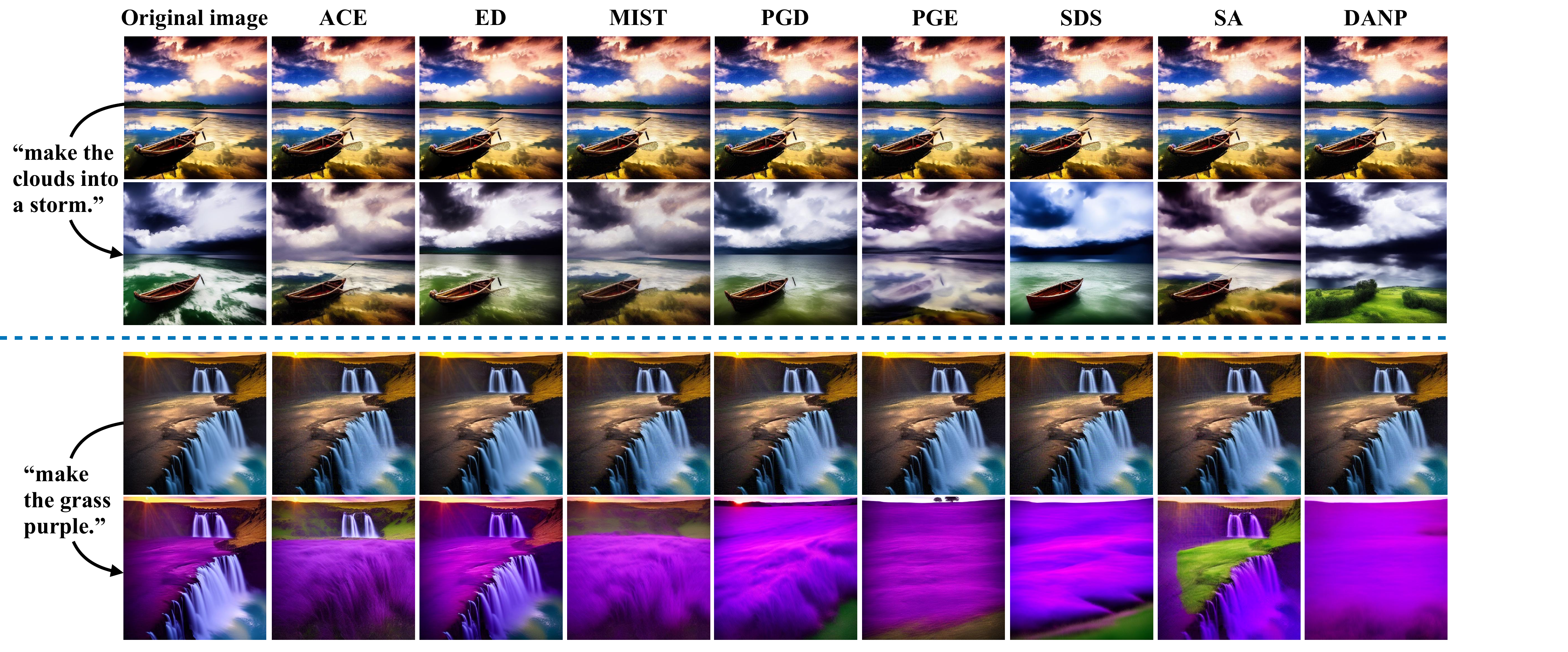}
    \caption{Qualitative comparison of the proposed DANP attack with previous image editing approaches on the \texttt{Instructpix2pix} model. Compared methods include ACE~\cite{Zheng2023TargetedAI}, ED~\cite{Chen2023EditShieldPU}, MIST~\cite{Liang2023MistTI}, PGD~\cite{Salman2023RaisingTC}, PGE~\cite{Salman2023RaisingTC}, SDS~\cite{Xue2023TowardEP} and SA~\cite{Lo_2024_CVPR}.}
    \label{fig:ins_visual}
\end{figure*}

\paragraph{Performance on Instructpix2pix}
The efficacy of DANP is first rigorously evaluated on the widely adopted \texttt{Instructpix2pix} model. The comprehensive results presented in Table~\ref{tab:quantitative_comparison_fid} unequivocally demonstrate that our method establishes a new state-of-the-art in image immunization. Across all six evaluation metrics, DANP consistently outperforms the seven competing baselines in both the original prompt and the more challenging unseen prompt scenarios, underscoring its superior effectiveness and robustness.

Under the original prompt condition, where the immunization is optimized for a known editing instruction, DANP shows a remarkable ability to disrupt the editing process. It achieves the highest dissimilarity scores in perceptual metric, registering an LPIPS of 0.4861. Concurrently, DANP obtains the lowest scores in all similarity-based metrics, indicating its potent defense capabilities against targeted edits.

To assess the generalization of our defense, we evaluate all methods against unseen prompts, a scenario that simulates real world applications where the exact editing instruction is unknown. In this stringent test, DANP continues to surpass all other approaches. It achieves a LPIPS score of 0.4651, a notable improvement over the existing best method SDS. The superiority of DANP is further affirmed by its leading performance in the similarity metrics, achieving the lowest PSNR, SSIM, VIFp, and FSIM values. This consistent performance highlights DANP's strong generalization and its ability to provide robust protection against a wide range of unforeseen malicious edits.

Furthermore, the superiority of DANP is vividly illustrated by the qualitative results presented in Fig.~\ref{fig:ins_visual}, which showcase the unique nature of its defensive mechanism. In the first example, with the prompt to ``make the clouds into a storm'', most baselines yield results that, while dissimilar to the target edit, still preserve the core semantic elements such as the stormy weather and the original boat. In stark contrast, the image immunized by our DANP leads to a much more profound editing failure. While the model does attempt to generate a storm, it completely misinterprets the scene's content, transforming the original sea into a field of grass and causing the boat to vanish entirely. An even more dramatic outcome is observed in the second example, which uses the prompt ``make the grass purple''. The edit applied to the image protected by our DANP method results in a catastrophic generative failure, where the entire image becomes saturated with an overwhelming purple hue instead of localizing the change to the grass. These visual comparisons provide compelling evidence that DANP induces a more comprehensive and destructive collapse of the editing process, far exceeding the partial disruptions achieved by prior works.

\begin{table*}[t!]
\centering
\caption{Quantitative performance comparison of various immunization methods on the \texttt{StableDiffusion-v1-4} model, evaluated under both original and unseen prompt scenarios.}
\label{tab:quantitative_comparison_fid_sd14}
\resizebox{\textwidth}{!}{%
\begin{tabular}{l ccccc ccccc }
\toprule
Method & \multicolumn{5}{c}{Original Prompt} & \multicolumn{5}{c}{Unseen Prompts} \\
\cmidrule(lr){2-6} \cmidrule(lr){7-11}
& PSNR $\downarrow$ & SSIM $\downarrow$ & LPIPS $\uparrow$ & VIFp $\downarrow$ & FSIM $\downarrow$ & PSNR $\downarrow$ & SSIM $\downarrow$ & LPIPS $\uparrow$ & VIFp $\downarrow$ & FSIM $\downarrow$ \\
\midrule
ACE~\cite{Zheng2023TargetedAI}   & 15.83 & 0.3567 & 0.5549 & 0.0530 & 0.6793 & 16.35 & 0.4115 & 0.5093 & 0.0669 & 0.7055 \\
SDS~\cite{Xue2023TowardEP}   & 15.96 & 0.3081 & 0.5473 & 0.0508 & 0.6812 & 16.13 & 0.3503 & 0.5275 & 0.0562 & 0.6900 \\
ED~\cite{Chen2023EditShieldPU}    & 16.52 & 0.4459 & 0.4700 & 0.0867 & 0.7236 & 16.55 & 0.4502 & 0.4670 & 0.0877 & 0.7229 \\
MIST~\cite{Liang2023MistTI}  & 15.63 & 0.3021 & 0.5593 & 0.0474 & 0.6682 & 16.01 & 0.3644 & 0.5292 & 0.0569 & 0.6905 \\
PGD~\cite{Salman2023RaisingTC}   & 15.79 & 0.3293 & 0.5474 & 0.0515 & 0.6779 & 16.13 & 0.3820 & 0.5189 & 0.0605 & 0.6961 \\
PGE~\cite{Salman2023RaisingTC}   & 15.97 & 0.2571 & 0.5817 & 0.0499 & 0.7093 & 16.32 & 0.3650 & 0.5141 & 0.0603 & 0.7111 \\
SA~\cite{Lo_2024_CVPR}    & 15.72 & 0.2803 & 0.5793 & 0.0412 & 0.6557 & 16.00 & 0.3584 & 0.5316 & 0.0547 & 0.6889 \\
\textbf{DANP} & \textbf{14.63} & \textbf{0.2378} & \textbf{0.6215} & \textbf{0.0350} & \textbf{0.6352} & \textbf{15.53} & \textbf{0.3419} & \textbf{0.5493} & \textbf{0.0514} & \textbf{0.6806} \\
\bottomrule
\end{tabular}%
}
\end{table*}

\begin{figure*}[t]
    \centering
    \includegraphics[width=1\linewidth]{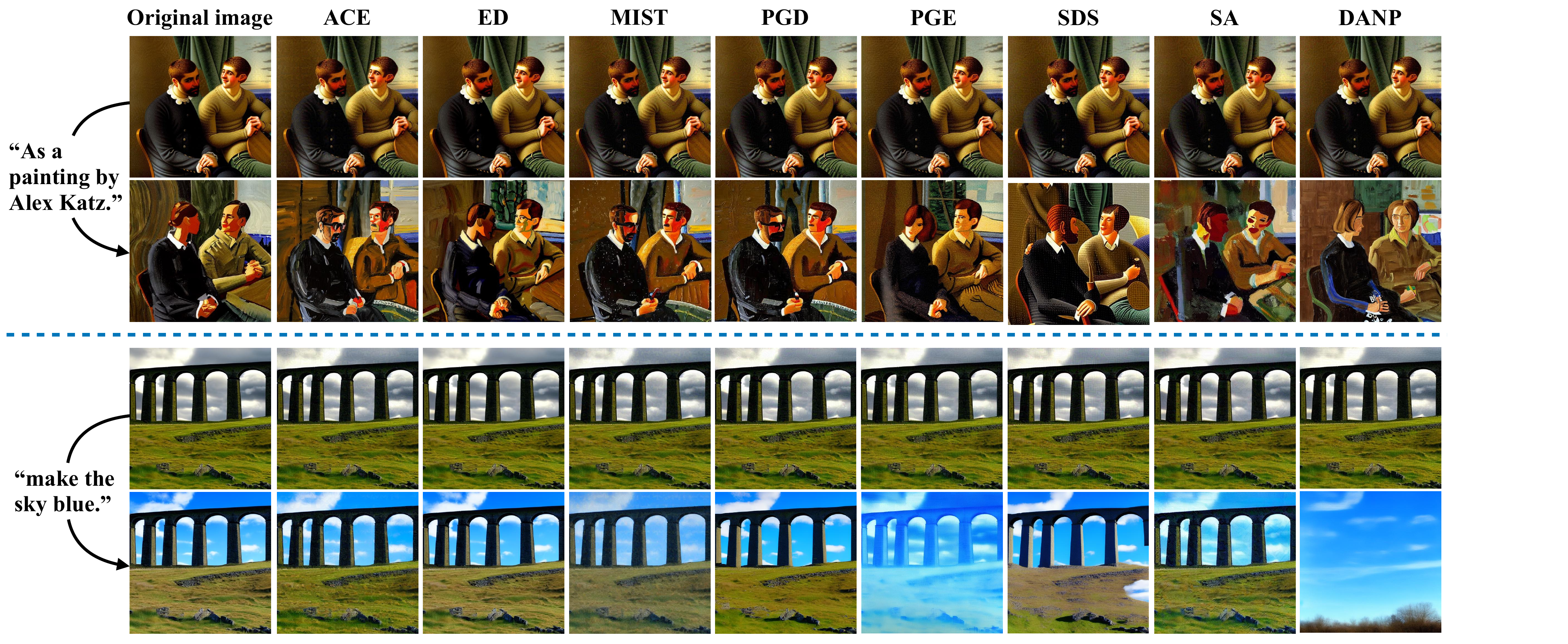}
    \caption{Qualitative comparison of the proposed DANP attack with previous image editing approaches on the \texttt{StableDiffusion-v1-4} model. Compared methods include ACE~\cite{Zheng2023TargetedAI}, ED~\cite{Chen2023EditShieldPU}, MIST~\cite{Liang2023MistTI}, PGD~\cite{Salman2023RaisingTC}, PGE~\cite{Salman2023RaisingTC}, SDS~\cite{Xue2023TowardEP} and SA~\cite{Lo_2024_CVPR}.}
    \label{fig:sd14_visual}
\end{figure*}

\paragraph{Performance on StableDiffusion-v1-4}
To further validate the generalizability of our approach, we extend our evaluation to the foundational \texttt{StableDiffusion-v1-4} model. The comprehensive results, summarized in Table~\ref{tab:quantitative_comparison_fid_sd14}, reaffirm the superiority of DANP, confirming that it maintains its state-of-the-art performance over all baseline methods on this distinct architecture. In the original prompt scenario, our method achieves a LPIPS score of 0.6215, significantly outperforming all existing methods. This substantial lead is maintained and even widened in the more demanding unseen prompt setting, where DANP secures a LPIPS score of 0.5493, surpassing the second-place score of 0.5316 from SA. This consistent dominance across all quantitative metrics underscores a powerful defense mechanism of our DANP. This quantitative success is mirrored in the qualitative outcomes shown in Fig.~\ref{fig:sd14_visual}, which reveal that DANP induces more severe editing failures than other methods. For instance, when prompted to ``make the sky blue'', our method does not merely inhibit the color change but triggers a complete generative collapse. This results in the model incorrectly applying a blue tint uniformly across the entire scene, including the bridge and landscape, indicating a failure in spatial and semantic localization. Furthermore, when defending against the stylistic prompt ``As a painting by Alex Katz'', DANP's immunization leads to a severe corruption of the model's core semantic understanding. Not only does it resist the style transfer, but it also introduces significant semantic errors, such as altering the gender of the two male subjects to female. These consistent and dominant results on a second major architecture underscore the effectiveness of our DANP for image editing immunization.

\begin{table*}[t!]
\centering
\caption{Quantitative performance comparison of various immunization methods on the \texttt{HQ-Edit} model, evaluated under both original and unseen prompt scenarios.}
\label{tab:quantitative_comparison_fid_hq}
\resizebox{\textwidth}{!}{%
\begin{tabular}{l ccccc ccccc }
\toprule
Method & \multicolumn{5}{c}{Original Prompt} & \multicolumn{5}{c}{Unseen Prompts} \\
\cmidrule(lr){2-6} \cmidrule(lr){7-11}
& PSNR $\downarrow$ & SSIM $\downarrow$ & LPIPS $\uparrow$ & VIFp $\downarrow$ & FSIM $\downarrow$ & PSNR $\downarrow$ & SSIM $\downarrow$ & LPIPS $\uparrow$ & VIFp $\downarrow$ & FSIM $\downarrow$ \\
\midrule
ACE~\cite{Zheng2023TargetedAI}   & 9.43 & 0.2990 & 0.6638 & 0.0446 & 0.5958 & 9.32 & 0.2856 & 0.6649 & 0.0426 & 0.5845 \\
SDS~\cite{Xue2023TowardEP}   & 8.99 & 0.2683 & 0.6703 & 0.0397 & 0.5816 & 9.29 & 0.2795 & 0.6705 & 0.0420 & 0.5839 \\
ED~\cite{Chen2023EditShieldPU}    & 9.92 & 0.3387 & 0.6075 & 0.0618 & 0.6155 & 9.17 & 0.2766 & 0.6672 & 0.0420 & 0.5798 \\
MIST~\cite{Liang2023MistTI}  & 9.71 & 0.3275 & 0.6692 & 0.0513 & 0.6016 & 9.33 & 0.2962 & 0.6534 & 0.0475 & 0.5869 \\
PGD~\cite{Salman2023RaisingTC}   & 9.42 & 0.2936 & 0.6609 & 0.0443 & 0.5917 & 9.30 & 0.2905 & 0.6641 & 0.0443 & 0.5848 \\
PGE~\cite{Salman2023RaisingTC}   & 9.06 & 0.2642 & 0.6769 & 0.0417 & 0.5855 & 9.29 & 0.2807 & 0.6655 & 0.0422 & 0.5826 \\
SA~\cite{Lo_2024_CVPR}    & 9.05 & 0.2699 & 0.6963 & 0.0404 & 0.5824 & 9.25 & 0.2783 & 0.6660 & 0.0422 & 0.5828 \\
\textbf{DANP} & \textbf{8.77} & \textbf{0.2564} & \textbf{0.7165} & \textbf{0.0386} & \textbf{0.5722} & \textbf{9.10} & \textbf{0.2695} & \textbf{0.6849} & \textbf{0.0389} & \textbf{0.5790} \\
\bottomrule
\end{tabular}%
}
\end{table*}

\begin{figure*}[t]
    \centering
    \includegraphics[width=1\linewidth]{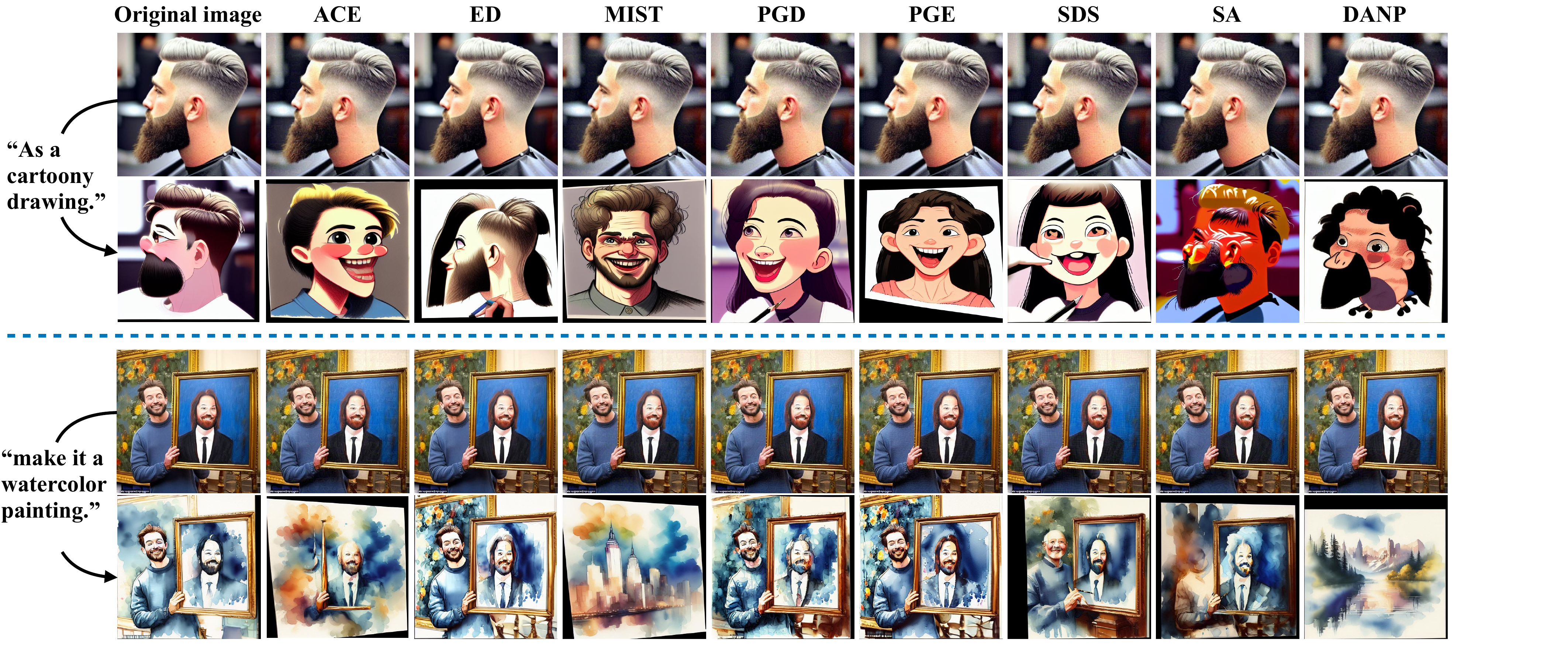}
    \caption{Qualitative comparison of the proposed DANP attack with previous image editing approaches on the \texttt{HQ-Edit} model. Compared methods include ACE~\cite{Zheng2023TargetedAI}, ED~\cite{Chen2023EditShieldPU}, MIST~\cite{Liang2023MistTI}, PGD~\cite{Salman2023RaisingTC}, PGE~\cite{Salman2023RaisingTC}, SDS~\cite{Xue2023TowardEP} and SA~\cite{Lo_2024_CVPR}.}
    \label{fig:hq_visual}
\end{figure*}

\paragraph{Performance on HQ-Edit}
Finally, we assess our method on the \texttt{HQ-Edit} model, a specialized high-resolution editor, to evaluate its performance in a different model paradigm. An interesting phenomenon emerges from the results presented in Table~\ref{tab:quantitative_comparison_fid_hq}. All tested immunization methods achieve remarkably strong defense scores, indicated by universally low PSNR values and high LPIPS scores. This suggests that the \texttt{HQ-Edit} model's architecture may be inherently more susceptible to adversarial immunization attacks compared to the other models. Even within this context of heightened overall defense effectiveness, where improvements are harder to achieve, DANP remains the strongest defense. It consistently ranks first across all metrics, achieving a leading LPIPS score of 0.7165 on original prompts and 0.6849 on unseen prompts. This consistent outperformance, though more marginal, demonstrates that DANP exhibits a discernible advantage even when all defenses are generally effective.

The qualitative results in Fig.~\ref{fig:hq_visual} further verify this conclusion. When prompted for a ``cartoony drawing,'' most methods produce recognizable cartoon styles. DANP, however, induces a more severe failure, generating a distorted and semantically incoherent image. A similar effect is seen with the ``watercolor painting'' prompt, where our method results in a chaotic and structurally broken image rather than a simple textural overlay. This shows that DANP excels at pushing the model beyond a failed edit into a state of complete generative collapse. Therefore, the evaluation on \texttt{HQ-Edit} confirms that while some models are easier to disrupt, DANP remains the most robust solution, providing the highest level of security across diverse model architectures.

\subsection{Analysis of Perturbation Imperceptibility}
A fundamental challenge in image immunization is balancing robust protection with visual fidelity. The ideal defense must add a perturbation that is potent enough to thwart edits yet subtle enough to be imperceptible. To assess this balance, we quantitatively evaluate the imperceptibility of all methods by measuring the similarity between the original source images and the final immunized versions. The comprehensive results, consolidated across all three models in Table~\ref{tab:imperceptibility_all_horizontal}, provide a clear view of this trade-off.

Specifically, the method ED achieves near-perfect imperceptibility (PSNR $> 46$), aligning with its optimization objective. However, this comes at a substantial cost to defensive efficacy, as demonstrated in Tab~\ref{tab:quantitative_comparison_fid}. DANP addresses this trade-off by achieving a superior balance: it maintains high imperceptibility with visual quality metrics comparable to leading defenses. On the Instructpix2pix dataset, its PSNR of 34.76 and SSIM of 0.8903 are competitive with top methods like ACE and SA. Crucially, DANP delivers this visual integrity alongside the state-of-the-art immunization performance. This demonstrates DANP's ability to provide best-in-class protection without introducing perceptible artifacts or compromising source image quality.

\begin{table*}[t!]
\centering
\caption{Quantitative analysis of perturbation imperceptibility. We compare the immunized images with the original source images across three different models.}
\label{tab:imperceptibility_all_horizontal}
\resizebox{\textwidth}{!}{%
\begin{tabular}{l ccccc ccccc ccccc}
\toprule
\multirow{2}{*}{Method} & \multicolumn{5}{c}{Instructpix2pix} & \multicolumn{5}{c}{HQ-Edit} & \multicolumn{5}{c}{StableDiffusion-v1-4} \\
\cmidrule(lr){2-6} \cmidrule(lr){7-11} \cmidrule(lr){12-16}
& PSNR$\uparrow$ & SSIM$\uparrow$ & LPIPS$\downarrow$ & VIFp$\uparrow$ & FSIM$\uparrow$ & PSNR$\uparrow$ & SSIM$\uparrow$ & LPIPS$\downarrow$ & VIFp$\uparrow$ & FSIM$\uparrow$ & PSNR$\uparrow$ & SSIM$\uparrow$ & LPIPS$\downarrow$ & VIFp$\uparrow$ & FSIM$\uparrow$ \\
\midrule
ACE~\cite{Zheng2023TargetedAI}  & 34.90 & 0.9020 & 0.2489 & 0.5608 & 0.9704 & 34.80 & 0.9001 & 0.2548 & 0.5576 & 0.9696 & 34.88 & 0.9051 & 0.2340 & 0.5660 & 0.9712 \\
SDS~\cite{Xue2023TowardEP}  & 34.44 & 0.8770 & 0.2594 & 0.5398 & 0.9651 & 34.40 & 0.8764 & 0.2629 & 0.5394 & 0.9657 & 34.75 & 0.8875 & 0.2479 & 0.5549 & 0.9686 \\
ED~\cite{Chen2023EditShieldPU}   & 47.02 & 0.9949 & 0.0106 & 0.9736 & 0.9990 & 46.54 & 0.9719 & 0.0118 & 0.9804 & 0.9992 & 47.25 & 0.9871 & 0.0099 & 0.9723 & 0.9994 \\
MIST~\cite{Liang2023MistTI} & 34.39 & 0.8888 & 0.2662 & 0.5474 & 0.9713 & 34.38 & 0.8884 & 0.2668 & 0.5470 & 0.9712 & 35.49 & 0.8957 & 0.2642 & 0.5840 & 0.9741 \\
PGD~\cite{Salman2023RaisingTC}  & 36.06 & 0.9112 & 0.2498 & 0.6014 & 0.9749 & 36.11 & 0.9114 & 0.2489 & 0.6037 & 0.9755 & 36.74 & 0.9200 & 0.2382 & 0.6277 & 0.9777 \\
PGE~\cite{Salman2023RaisingTC}  & 34.54 & 0.8787 & 0.2360 & 0.5343 & 0.9729 & 35.23 & 0.8814 & 0.2335 & 0.5384 & 0.9813 & 34.98 & 0.8804 & 0.2307 & 0.5444 & 0.9762 \\
SA~\cite{Lo_2024_CVPR}   & 34.82 & 0.8915 & 0.2699 & 0.5547 & 0.9701 & 35.20 & 0.8991 & 0.2438 & 0.5778 & 0.9755 & 34.60 & 0.8866 & 0.2674 & 0.5582 & 0.9718 \\
DANP & 34.76 & 0.8903 & 0.2740 & 0.5548 & 0.9680 & 35.16 & 0.8941 & 0.2671 & 0.5748 & 0.9769 & 34.81 & 0.8678 & 0.2914 & 0.5275 & 0.9709 \\
\bottomrule
\end{tabular}%
}
\end{table*}

\begin{table*}[t!]
\centering
\caption{Ablation study on the DAA and NBA components of DANP.}
\label{tab:ablation_main}
\resizebox{\textwidth}{!}{%
\begin{tabular}{l ccccc ccccc }
\toprule
Method & \multicolumn{5}{c}{Original Prompt} & \multicolumn{5}{c}{Unseen Prompt} \\
\cmidrule(lr){2-6} \cmidrule(lr){7-11}
& PSNR $\downarrow$ & SSIM $\downarrow$ & LPIPS $\uparrow$ & VIFp $\downarrow$ & FSIM $\downarrow$ & PSNR $\downarrow$ & SSIM $\downarrow$ & LPIPS $\uparrow$ & VIFp $\downarrow$ & FSIM $\downarrow$ \\
\midrule
w/o DAA & 16.34 & 0.6135 & 0.4508 & 0.2192 & 0.8135 & 16.28 & 0.5789 & 0.4303 & 0.1862 & 0.7818 \\
w/o NBA & 15.07 & 0.5502 & 0.4817 & 0.1793 & 0.7644 & 15.45 & 0.5515 & 0.4551 & 0.1610 & 0.7723 \\
DANP & 14.67 & 0.5487 & 0.4861 & 0.1764 & 0.7519 & 14.99 & 0.5498 & 0.4651 & 0.1520 & 0.7611 \\
\bottomrule
\end{tabular}%
}
\end{table*}

\begin{table*}[t!]
\centering
\caption{Ablation study on the number of bins ($L$) for Kapur's method. We report immunization performance across various metrics and the computation time per iteration. }
\label{tab:ablation_bins}
\begin{tabular}{r ccccc c}
\toprule
\multicolumn{1}{c}{Bins} & PSNR $\downarrow$ & SSIM $\downarrow$ & LPIPS $\uparrow$ & VIFp $\downarrow$ & FSIM $\downarrow$ & Time/iter (s) $\downarrow$ \\
\midrule
32   & \textbf{14.23} & 0.5570 & 0.4630 & 0.2009 & 0.7675 & 3.82 \\
64   & 14.54 & 0.5676 & 0.4750 & 0.1946 & 0.7910 & 4.34 \\
\textbf{128} & 14.67 & \textbf{0.5487} & \textbf{0.4861} & \textbf{0.1764} & \textbf{0.7519} & \textbf{4.16} \\
256  & 14.96 & 0.5664 & 0.4787 & 0.1835 & 0.7851 & 6.27 \\
512  & 15.01 & 0.5790 & 0.4800 & 0.1887 & 0.7898 & 8.73 \\
1024 & 15.14 & 0.5675 & 0.4807 & 0.1854 & 0.7892 & 13.51 \\
\bottomrule
\end{tabular}
\end{table*}

\begin{figure}[t!]
\centering
\includegraphics[width=\linewidth]{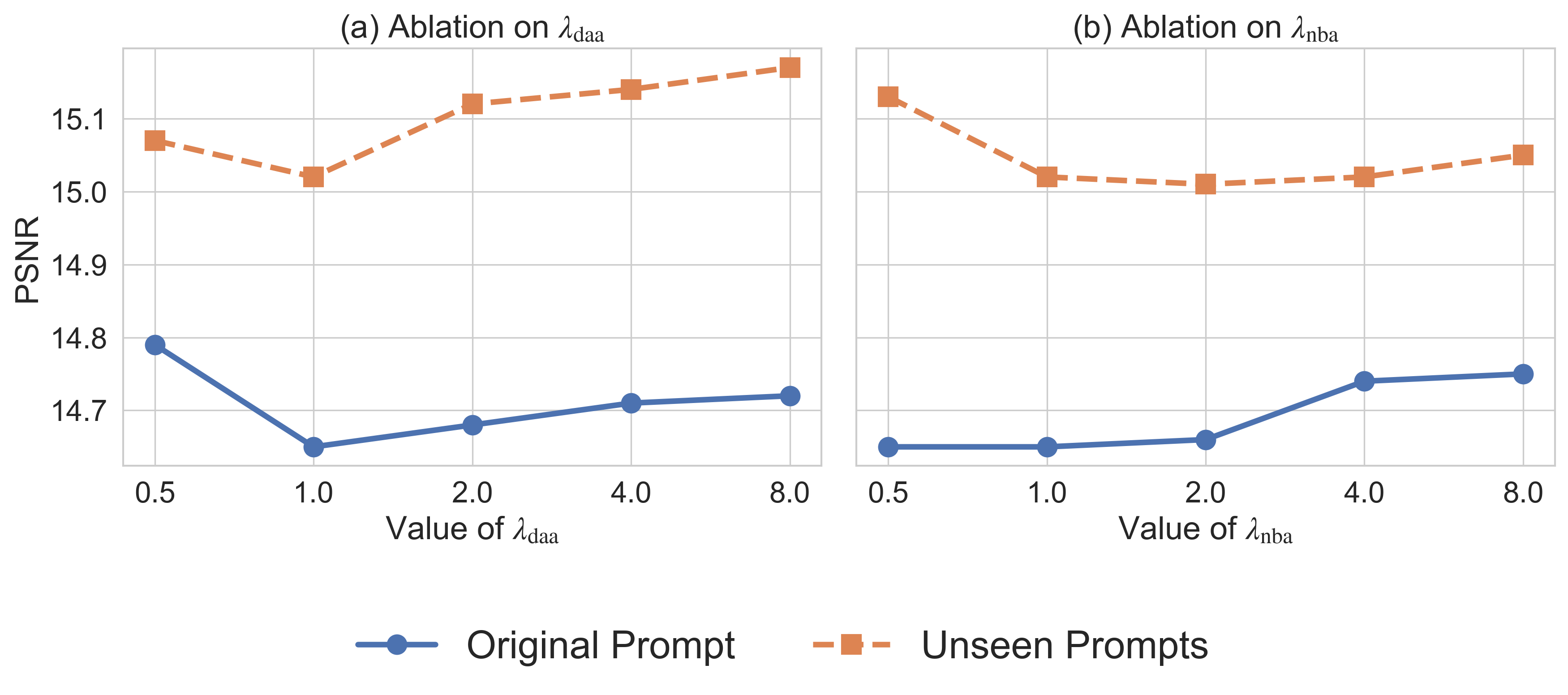}
\caption{Ablation study on the hyperparameters (a) $\lambda_{\text{daa}}$ and (b) $\lambda_{\text{nba}}$. The immunization performance, measured by PSNR (lower is better), remains highly stable across several orders of magnitude for both parameters.}
\label{fig:ablation_lambda}
\end{figure}
\subsection{Ablation Studies}
\paragraph{Effectiveness of DANP Components}
To validate the individual contributions and synergy of DANP's core components, \ie, Dual Attention guided Attack (DAA), and Noise Based Attack (NBA), we conduct a detailed ablation study on the editing model \texttt{Instructpix2pix}. We systematically deactivate each component (DAA or NBA) and evaluate performance. The results in Table~\ref{tab:ablation_main} confirm that both are indispensable, and their combination is critical for peak  performance.

First, disabling DAA (the w/o DAA variant, relying solely on NBA) yields moderate performance (e.g., LPIPS = 0.4508 on original prompts). Integrating DAA to form the complete DANP, however, substantially improves performance, increasing the LPIPS to 0.4861. This significant gain underscores DAA's critical role: by disrupting the model's cross-attention, it serves as the primary immunization mechanism, preventing correct localization of edit regions.

Next, we isolate the DAA component (\textit{w/o NBA} variant), which achieves a strong standalone LPIPS of 0.4817, demonstrating its inherent effectiveness. However, incorporating the NBA yields a significant synergistic boost, elevating the LPIPS to 0.4861 on original prompts and from 0.4551 to 0.4651 on unseen prompts. This synergy arises from their complementary mechanisms: the DAA disrupts the model's semantic understanding of where to edit, while the NBA corrupts the generative process governing how to render the edits. Attacking both these distinct, critical stages simultaneously creates a more robust and comprehensive defense than targeting either stage alone. Thus, the synergistic interplay between DAA and NBA is essential for DANP's superior efficacy.

\subsection{Analysis of Hyperparameters}
This section analyzes the impact of key hyperparameters on DANP's performance. First, we investigate the sensitivity of DANP to the loss weights $\lambda_{\text{daa}}$ and $\lambda_{\text{nba}}$. Notably, the DAA and NBA loss terms differ significantly in their order of magnitude. To ensure a fair and balanced ablation, we scale the NBA loss to a comparable level with the DAA loss. As illustrated in Fig.~\ref{fig:ablation_lambda}, the immunization performance exhibits remarkable stability even when these parameters are varied across a significant range from 0.5 to 8. The resulting changes in the final PSNR scores are negligible for both original and unseen prompts. This low sensitivity demonstrates that DANP is robust and does not require extensive hyperparameter tuning to achieve strong results. Consequently, we set $\lambda_{\text{daa}} = 1.0$ and $\lambda_{\text{nba}} = 1.0$ as a simple and effective choice for all our experiments.

Second, we analyze the influence of the number of bins \texttt{L} used in Kapur's method for dynamic thresholding. The results presented in Table~\ref{tab:ablation_bins} reveal a clear trade-off between immunization effectiveness and computational cost. As \texttt{L} increases from 32 to 128, key performance metric like LPIPS improve significantly, reaching their optimal values at \texttt{L=128} with a LPIPS score of 0.4861. However, further increasing \texttt{L} beyond 128 leads to a sharp rise in computation time per iteration with no corresponding benefit to immunization performance. Although the PSNR is lowest at \texttt{L=32}, this is outweighed by the superior performance across all other more comprehensive metrics at \texttt{L=128}. Therefore, we select \texttt{L=128} as the optimal setting, as it offers the best immunization performance at a reasonable computational cost.

\section{Conclusion}
In this paper, we propose the Dual Attention-Guided Noise Perturbation (DANP), a novel image immunization method designed to protect images from malicious editing by diffusion-based models. DANP employs a more comprehensive, dual-pronged strategy that simultaneously disrupts both the semantic understanding and the generative process of the editing model through its two synergistic components, \ie, the Dual Attention-guided Attack (DAA) and the Noise-Based Attack (NBA). Extensive experiments demonstrate that DANP achieves state-of-the-art performance across multiple editing models, outperforming existing methods and showing strong robustness against unseen editing prompts. By simultaneously compromising both \textit{where to edit} and \textit{how to edit}, DANP provides a more powerful and resilient solution for defending against malicious image manipulation.


\bibliographystyle{IEEEtran} 

\bibliography{myref}


 




\vfill

\end{document}